\documentclass{article}
\usepackage{ijcai25}

\usepackage{times}
\usepackage{soul}
\usepackage{url}
\usepackage[hidelinks]{hyperref}
\usepackage[utf8]{inputenc}
\usepackage[small]{caption}
\usepackage{graphicx}
\usepackage{amsmath}
\usepackage{amsthm}
\usepackage{booktabs}
\usepackage{algorithm}
\usepackage{algorithmic}
\usepackage[switch]{lineno}
\usepackage{multirow}
\usepackage{pifont}

\title{Incorporating Legal Logic into Deep Learning: An Intelligent
Approach to Probation Prediction}

\author{
Qinghua Wang\textsuperscript{\rm 1}\and
Xu Zhang\textsuperscript{\rm 1}\and
Lingyan Yang\textsuperscript{\rm 1}\and\\
Rui Shao\textsuperscript{\rm 1}\and
Bonan Wang\textsuperscript{\rm 2}\and
Fang Wang\textsuperscript{\rm 1}\and
Cunquan Qu\textsuperscript{\rm 1}
\\
\affiliations
$^1$Shandong University\\
$^2$University of Macau\\
\emails
\{qinghuawang, xuzhang, 202321227, 202420955\}@mail.sdu.edu.cn,
bonan.wang@connect.um.edu.mo,
\{wangfang226, cqqu\}@sdu.edu.cn
}

\begin{document}

\maketitle

\begin{abstract}
    Probation is a crucial institution in modern criminal law, embodying the principles of fairness and justice while contributing to the harmonious development of society. Despite its importance, the current Intelligent Judicial Assistant System (IJAS) lacks dedicated methods for probation prediction, and research on the underlying factors influencing probation eligibility remains limited. In addition, probation eligibility requires a comprehensive analysis of both criminal circumstances and remorse. Much of the existing research in IJAS relies primarily on data-driven methodologies, which often overlooks the legal logic underpinning judicial decision-making. To address this gap, we propose a novel approach that integrates legal logic into deep learning models for probation prediction, implemented in three distinct stages. First, we construct a specialized probation dataset that includes fact descriptions and probation legal elements (PLEs). Second, we design a distinct probation prediction model named the Multi-Task Dual-Theory Probation Prediction Model (MT-DT), which is grounded in the legal logic of probation and the \textit{Dual-Track Theory of Punishment}. Finally, our experiments on the probation dataset demonstrate that the MT-DT model outperforms baseline models, and an analysis of the underlying legal logic further validates the effectiveness of the proposed approach.
\end{abstract}

\section{Introduction}

In recent years, the application of artificial intelligence (AI) in the Intelligent Judicial Assistant System (IJAS) has increased steadily \cite{ref19,ref38,ref39,ref40}, primarily with the aim of improving judicial efficiency. Research has mainly focused on improving the performance of AI from two perspectives. The first focuses on continuously refining the extraction and learning of key information from judgment documents \cite{ref1,ref20,ref21}. These methods, however, mainly concentrate on superficial information and fail to deeply analyze the legal interpretation of texts. The second adopts data-driven techniques to build effective neural network models for legal judgment prediction (LJP) \cite{ref2,ref3,ref4}. However, these approaches often overlook the inherent legal logic. 
\begin{figure}[!t]
    \centering
    \includegraphics[width=\columnwidth]{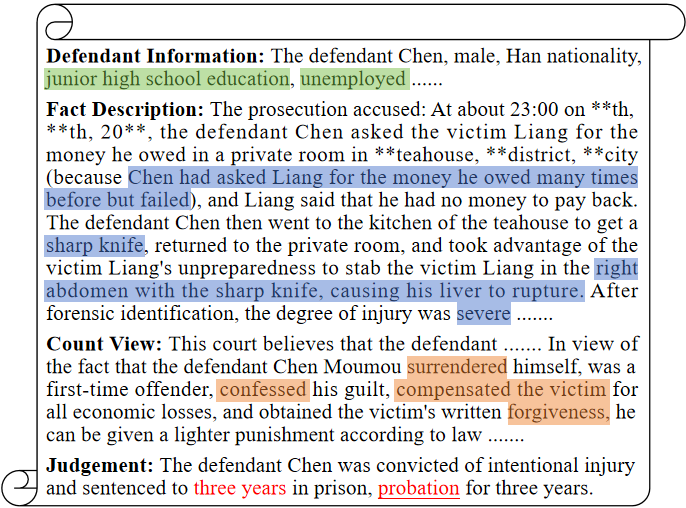}
    \caption{Illustrative example of information related to probation conditions. Highlight: information related to probation conditions. Red text: the prerequisite of probation. Red underlined text: probation eligibility.}%
    \label{fig:headfigure}
\end{figure}

Probation is a crucial component of sentencing \cite{ref41,ref42,ref43}. On the one hand, it offers criminals an opportunity for rehabilitation through non-custodial measures, enabling them to reform and reducing the likelihood of re-offending \cite{ref17,ref44,ref45}. On the other hand, probation helps alleviate the burden on the prison system and contributes to social harmony by facilitating the reintegration of criminal \cite{ref18,ref46,ref47}. Despite its importance, current IJAS lack dedicated methods for probation prediction, and research on the underlying factors influencing probation decisions remains limited. 

Article 72 of the {\it Criminal Law of the People's Republic of China\footnote{\url{http://xingfa.org/}}} ({\it Art. 72})  stipulates that criminals sentenced to detention or fixed-term imprisonment of three years or less may be granted probation if they meet the following conditions at the same time: 
\begin{itemize}
    \item \textbf{Condition (a).} Mild circumstances of the crime.
    \item \textbf{Condition (b).} Evidence of remorse.
    \item \textbf{Condition (c).} No danger of re-offending.
    \item \textbf{Condition (d).} No significant negative impact on the community.
\end{itemize}

In addition, criminals sentenced to detention or fixed-term imprisonment of 3 years or less must be granted probation if they are under 18 years old, pregnant, or over 75 years old. In the Chinese judicial system, both the prerequisite and substantive conditions for probation form a crucial theoretical foundation for determining probation eligibility, which guide judges to assess a criminal's probation eligibility. Figure~\ref{fig:headfigure} illustrates a case example of intentional injury, highlighting key information relevant to probation eligibility of criminal. In this case, the defendant Chen was sentenced to a three-year prison term, meeting the prerequisites for probation. In addition, his evidence of remorse and a low risk of recidivism further support his probation eligibility. 


Due to limited AI adoption in probation research, we aim to enhance deep learning models with legal logic for more accurate predictions. However, this poses several challenges:

\paragraph{Diverse Legal Information in LJP Tasks.} Different prediction tasks focus on distinct types of legal information. For example, in charge prediction tasks \cite{ref23,ref48,zhang2024hd}, it is essential to incorporate not only the fact description but also the specific elements that constitute each charge, enabling a deeper understanding of its legal implications. In case matching tasks \cite{ref22,ref49}, the relevant legal articles and the semantic relationships among entities must be integrated for similarity matching. Similarly, in probation prediction tasks, the PLEs play a particularly crucial role. Therefore, combining fact description with PLEs is critical for improving the performance of probation prediction models.

\begin{figure*}[!t]
\centering
\includegraphics[width=0.9\textwidth]{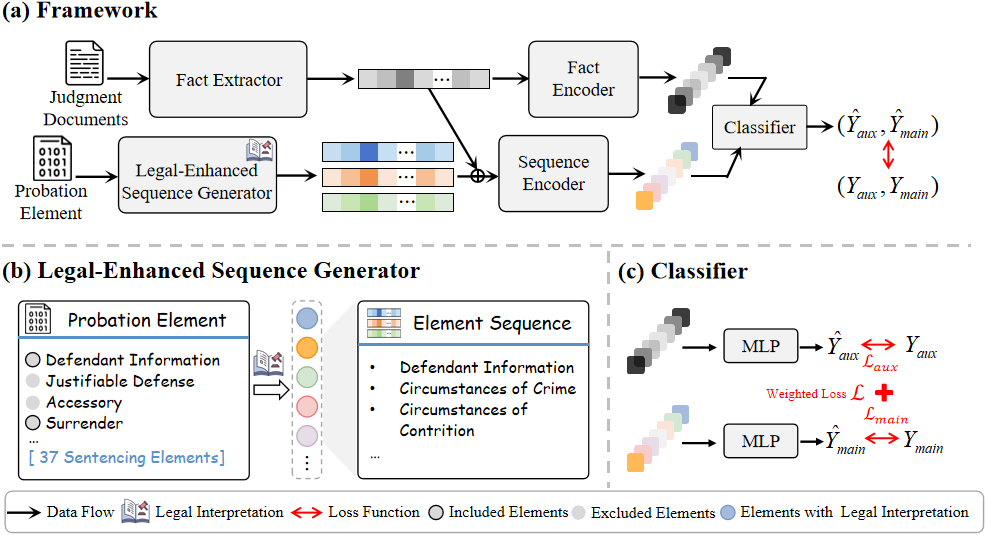}  
\caption{The overall architecture of the proposed MT-DT. It consists of three main steps: (1) Data Processing. This step includes the Fact Extractor and the Legal-Enhanced Sequence Generator. (2) Text Encoder. This step mainly encodes the text to obtain their vector representation. (3) Multi-Task Classifier. This module simulates the probation decision-making process and predicts probation eligibility.}
\label{fig:framework}
\end{figure*}

\paragraph{Legal Logic in Multi-Task Learning.} Multi-task learning approaches \cite{ref16} have been employed to predict multiple labels simultaneously. However, these methods typically share the same input across all tasks, neglecting the logical dependencies that exist between tasks. However, in probation prediction task, different information must be provided for each subtask, based on the specific legal logic underlying probation decisions. This task-specific approach aligns with legal logic and has the potential to significantly enhance both the accuracy and interpretability of the predictions.

To address the first challenge, we construct a dataset tailored for probation tasks, focusing on the crime of intentional injury as a case study. This crime exemplifies the conflict between its inherently high subjective malice and the "minor circumstances" required for probation eligibility. To navigate this conflict in probation cases involving intentional injury, we begin by analyzing the interpretation of "minor circumstances" within the substantive conditions of probation. Currently, two mainstream viewpoints exist. First, the crime must involve relatively minor circumstances, indicating a low degree of personal danger posed by the criminal, thereby fulfilling the retribution purpose of probation, referred to as the "retributive factor". Second, the minor circumstances of the crime also suggest a low risk of re-offending, which satisfy the preventive purpose of probation, referred to as the "preventive factor". In this study, we extract the fact description to represent the retributive factors. Additionally, guided by legal expertise, we obtain PLEs reflecting preventive factors and other substantive conditions.

To address the second challenge, we first propose the \textbf{T}wo-\textbf{S}tage Probation Prediction Model Based on P\textbf{LE}s (TS-LE), which focuses on preventive factors using a cascading prediction framework. Building upon TS-LE, we introduce the \textbf{T}wo-\textbf{S}tage \textbf{D}ual-\textbf{T}heory Probation Prediction Model (TS-DT), which integrates both retributive and preventive factors, aligning with the \textit{Dual-Track Theory of Punishment} to achieve a more balanced decision-making approach. Finally, extending TS-DT, the \textbf{M}ulti-\textbf{T}ask \textbf{D}ual-\textbf{T}heory Probation Prediction Model (MT-DT) adopts a multi-task framework, treating the prediction of prerequisites as an auxiliary task. This enhancement allows for the simultaneous optimization of prerequisites and substantive conditions, further improving prediction accuracy and interpretability. These models incorporate legal interpretation encoding of PLEs to better simulate the judicial decision-making process. By combining fact description with PLEs, these models provide a comprehensive and systematic assessment of probation eligibility. Validation on our constructed dataset demonstrates their effectiveness, with each successive model offering improved performance and deeper insights into the legal logic underlying probation decisions.

To sum up, this paper has three main contributions, summarized as follows:
\begin{itemize}
    \item \textbf{Dataset Construction.} We construct a probation prediction dataset, which includes fact description and PLEs with legal interpretations.
    \item \textbf{Legal Logic Analysis and Model Development.} We conduct an in-depth analysis of the legal logic behind probation eligibility and proposed three progressively advanced models, TS-LE, TS-DT and MT-DT, to simulate the decision-making process for probation.
    \item \textbf{Model Validation and Insights.} Experimental results on the constructed dataset show that the MT-DT model effectively integrates both the retributive and preventive factors, while also addressing the "error amplification" problem in cascading tasks.
\end{itemize}

\section{Related Work}

\subsection{Exploration of Factors in Probation} Several studies have explored the factors influencing probation eligibility. Cordier {\it et al.} \cite{ref12} analyzed the link between legal measures and recidivism risk, offering insights into probation outcomes. Similarly, Sims {\it et al.} \cite{ref13} studied 2,850 felony probationers in North Carolina and identified key factors such as demographics, criminal history, and probation conditions. Furthermore, Louden {\it et al.} \cite{ref14} used simulation-based experiments to highlight the complex role of mental health issues in probation management. Their findings emphasized the need for customized support and diverse intervention strategies. Williams {\it et al.} \cite{ref15} developed a tool to assess recidivism risk among juvenile offenders. All of these approaches primarily rely on mathematical and statistical methods.

\subsection{Application of Multi-Task Models in LJP} Multi-task learning models have been applied to various LJP tasks. For instance, Lyu {\it et al.} \cite{ref25} proposed a method to enhance legal judgment prediction by reinforcing the extraction of criminal elements, integrating these elements into content inputs for multi-task predictors, with the goal of improving model accuracy. Yao {\it et al.} \cite{ref16} introduced a gated hierarchical multi-task learning network, which models different legal tasks at multiple granularities, thereby improving the predictive performance of judicial decisions. Additionally, Yue {\it et al.} \cite{ref26} proposed the NeurJudge, an environment-aware approach that divides fact description into different scenarios using intermediate subtask results, which are then leveraged to predict other subtasks. These studies provide important insights for our research on intelligent probation prediction models.

To the best of our knowledge, this is the first work to integrate legal logic with AI techniques for probation prediction. Our approach focuses on developing deep learning models that incorporate legal logic, with the aim of achieving efficient and reliable probation prediction.

\section{Method}
\subsection{Overview}
Figure~\ref{fig:framework} showcases the overall pipeline of our model—data Processing, text Encoder, and multi-task classifier—into a cohesive system.


\subsection{Preliminaries}
\paragraph{Probation Prediction.}
According to {\it Art. 72}, probation applies only to criminals sentenced to three years or less in prison or detention. Therefore, the probation prediction consists of two subtasks: \textbf{Task 1:} Predict whether the criminal's sentence meets the prerequisite of three years or less, or detention. \textbf{Task 2:} Predict the probation eligibility of criminal.

\paragraph{Retributive (Preventive) Factors.} By analyzing the substantive conditions for probation, \textbf{Condition (a)} incorporates retributive or preventive factors. Retributive factors refer to elements that embody the principle of retribution within the probation system, primarily focusing on the criminal's personal danger or culpability. In this study, we represent retributive factors using fact description. Preventive factors, on the other hand, reflect the preventive purpose of probation and are included in the PLEs.

\paragraph{The Dual (Single)-Track Theory of Punishment.} When \textbf{Condition (a)} accounts for the retribution factor, it is known as the {\it Dual-Track Theory of Punishment}. In contrast, when only the prevention factor is considered, it is known as the {\it Single-Track Theory of Punishment}. This distinction arises because the retribution factor has been incorporated into the sentencing process as a prerequisite for probation. Re-incorporating this factor would constitute double retribution.

\paragraph{Probation Legal Elements.} Guided by legal expertise, we analyze the substantive conditions for probation eligibility as outlined in {\it Art. 72}, as follows:
\begin{itemize}
    \item \textbf{Mild Circumstances of the Crime.} For retribution, the fact description includes numerous retributive factors, such as the criminal's modus operandi and the consequences caused. For prevention, the underlying causes are considered, along with other subjective and objective circumstances, such as whether the criminal was forced to commit the crime. 

    \item \textbf{Evidence of Remorse.} According to Article 225 of the {\it Interpretation on the Application of the Criminal Procedure Law of the People's Republic of China}, remorse requires a comprehensive assessment of the criminal's post-crime behavior, including actions such as compensation, expressions of remorse, and whether forgiveness has been obtained.

    \item \textbf{No Danger of Re-offending.} This condition should be analyzed from two perspectives. One is the dangerousness of the criminal’s behavior, which focuses on the motivations behind the crime. Elements such as excessive self-defense (i.e., the criminal was coerced into committing the offense) are indicative of a lower risk of recidivism. The other perspective is the dangerousness of the individual, which considers the criminal history, as well as personal factors such as age, occupation, education, family environment, and social circumstances.

    \item \textbf{No Significant Negative Impact on the Community.} This condition consists of a variety of factors, including the criminal’s motives, methods, and the outcomes of the crime, to determine the extent of social harm caused by the offense.
\end{itemize}

\begin{figure}[!t]
\centering
\includegraphics[width=0.97\linewidth]{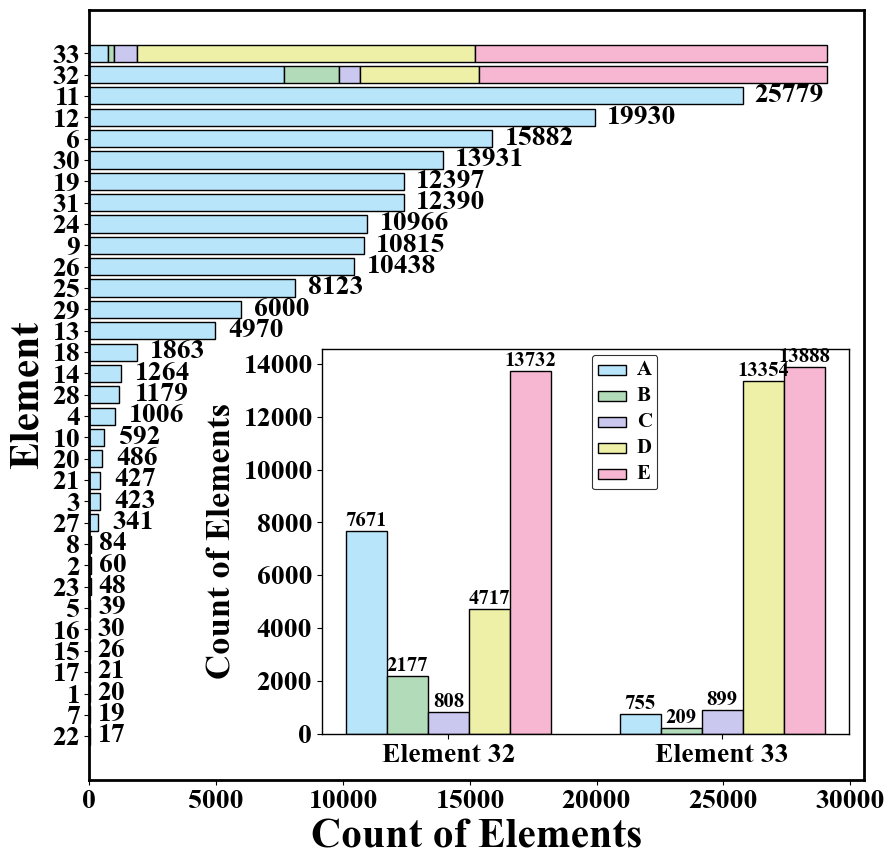}  
\caption{ Distribution of frequencies for each element in the dataset, where elements 32 and 33 each contain five distinct values.}
\label{fig:elements}
\end{figure}

Based on the analysis of the substantive conditions, we identify 33 distinct probation legal elements. 
The distribution of their frequencies is presented in Figure~\ref{fig:elements}. 

\subsection{Problem Formulation}
We assume that the fact description is denoted as \( F \) and the text sequence of PLEs as \( Q \). Based on \( F \) and \( Q \), the probation prediction task aims to learn a function \( \phi \), where \( \hat{y} = \phi (F, Q) \), to predict the probation eligibility of the criminal. This task consists of two subtasks: \( T = \{ T_{\text{aux}}, T_{\text{main}} \} \). The corresponding prediction results are \( \hat{y}_{\text{aux}} \in \{ 0, 1 \} \) and \( \hat{y}_{\text{main}} \in \{ 0, 1 \} \).

\subsection{Legal-Enhanced Sequence Generator}
As illustrated in Figure~\ref{fig:framework} (b), to provide legal implication to PLMs, we obtain the interpretation of each element from the encyclopedia entries and generate a corresponding text sequence. More specifically, we first apply regularization to match the elements of each case, as follows:
\begin{equation}
    \mathcal{V}_i = \mathcal{H}(R, E), \quad i = 1, 2, \dots, N, 
\end{equation}
where $\mathcal{V}_i = \{v_{i_{1}}, v_{i_{2}}, \dots, v_{i_{k}}\}$ is a vector that represents the presence of elements in the $i$-th judgement document, and $v_{i_{k}} \in \{0, 1\}$ indicates whether the $k$-th element is present ($v_{i_{k}} = 1$) or absent ($v_{i_{k}} = 0$). Here, $\mathcal{H}$ is the extraction function, $R$ is the extraction rule, $E$ represents the set of PLEs, and $N$ is the number of judgement document.

Then, we extract the legal interpretation for each element from the encyclopedic entries, forming $\mathcal{A}$ as:
\begin{equation}
    \mathcal{A} = \{ A_{1}, A_{2}, \dots, A_{k} \},
\end{equation}
where \( A_k\) mains the legal meaning of the $k$-th element. 

Finally, we search the knowledge base for the legal interpretation corresponding to the elements in the case and generate a text sequence of elements unique to the case $Q_i$ as:
\begin{equation}
    Q_i = \mathcal{G} (\mathcal{A}, \mathcal{V}_i),
\end{equation}
where \( \mathcal{G}\) represents the generating function.  

\subsection{Training and Prediction}
In the multi-task training of MT-DT, the fact description is encoded as input for the auxiliary task, represented as:
\begin{equation}
    w_{\text{aux}} = \text{ENC}_\text{aux}(F),
\end{equation}
where $F$ denotes a sequence representing the fact description, \(F = \{f_1, f_2, \dots, f_m\}\), and $m$ is the length of sequence.

\begin{table}[!t]
  \centering
  \begin{minipage}[t]{\linewidth}
    \centering
    \captionsetup{labelformat=empty,skip=1pt}
    \setlength{\tabcolsep}{1.8mm}
    \resizebox{\linewidth}{!}{
      \begin{tabular}{cccc}
        \toprule 
        & Framework & Input Information & Legal Logic  \\  \midrule
        \multirow{2}{*}{TS-LE} & \multirow{2}{*}{\shortstack{Cascaded Two- \\ Stage Model}} & Task1: F & \multirow{2}{*}{\shortstack{Single-Track Theory \\ of Punishment}} \\
                       &                                                          & Task2: Q &                               \\ \midrule
        \multirow{2}{*}{TS-DT} & \multirow{2}{*}{\shortstack{Cascaded Two- \\ Stage Model}} & Task1: F & \multirow{2}{*}{\shortstack{Dual-Track Theory \\ of Punishment}} \\
                       &                                                          & Task2: F \& Q &                               \\ \midrule                    
        \multirow{2}{*}{MT-DT} & \multirow{2}{*}{Multi-task Model} & \multirow{2}{*}{F \& Q} & \multirow{2}{*}{\shortstack{Dual-Track Theory \\ of Punishment}} \\
        & & & \\ 
        \bottomrule     
        \end{tabular}
    }
    \caption*{(a) Models with varying legal logics and framework.}
  \end{minipage}
  
  
  \begin{minipage}[t]{\linewidth}
    \centering
    \captionsetup{labelformat=empty,skip=1pt}
    
    \setlength{\tabcolsep}{3mm}
    \resizebox{\linewidth}{!}{
     \begin{tabular}{cccccc}
        \toprule 
        Model & & ACC(\%) & MP(\%) & MR(\%) & F1(\%) \\ \midrule
        \multirow{2}{*}{TS-LE} & Task 1 & 95.17 & 83.55 & 86.77 & 85.06 \\
        & Task 2 & 69.65 & 68.16 & 67.12 & 66.95 \\ \midrule
        \multirow{2}{*}{TS-DT} & Task 1 & 95.17 & 83.55 & 86.77 & 85.06 \\
        & Task 2 & 72.88 & 70.28 & 73.13 & 72.45 \\ \midrule
        MT-DT & Task 2 & 88.15 & 84.01 & 87.57 & 85.47 \\
        \bottomrule  
        \end{tabular}
    }
    \caption*{(b) Results on TS-LE, TS-DT, and MT-DT}
  \end{minipage}
  
  \caption{Performance comparison of three legal logic enhanced probation prediction models.}
  \label{tab:threemodel}
\end{table} 

According to the {\it Dual-Track Theory of Punishment}, we concatenate the fact description and the text sequences of PLEs, encoding them jointly as input for the main task. Specifically, this process can be represented as follows:
\begin{equation}
    w_{\text{main}} = \text{ENC}_\text{main}(F,Q),
\end{equation}
where $Q$ denotes a word sequence representing the elements, \(Q = \{q_1, q_2, \dots, q_n\}\), and $n$ is the length of sequence.

To obtain prediction results for the auxiliary and main task, we use two MLP classifiers:
\begin{equation}
\begin{aligned}
    \hat{y}_{\text{aux}} = \text{softmax}(\text{MLP}(w_{\text{aux}})), \\
    \hat{y}_{\text{main}} = \text{softmax}(\text{MLP}(w_{\text{main}})). 
\end{aligned}
\end{equation}

In order to train the multi-task probation predictor, we compute the cross-entropy loss function for both the auxiliary and main task. The loss for the auxiliary task is formally computed as:
\begin{equation}
    \mathcal{L}_{aux} = - \sum_{c=1}^{N} y_{aux_c} \log(\hat{y}_{aux_c}),
\end{equation}
where $y_{aux_c}$ represents the ground truth for the auxiliary task prediction associated with the $c$-th judgment document. Similarly, the loss for the main task is formally computed as:
\begin{equation}
    \mathcal{L}_{main} = - \sum_{c=1}^{N} y_{main_c} \log(\hat{y}_{main_c}),
\end{equation}
where $y_{main_c}$ represents the ground truth for the main task prediction associated with the $c$-th judgment document.

During multi-task training, we sum the losses of the different subtasks as the total loss to train the multi-task model:

\begin{equation}
    \mathcal{L} = \mathcal{L}_{\text{main}} + \lambda\mathcal{L}_{aux},
\end{equation}
where \(\lambda\) represents the weight of the auxiliary task loss.

\section{Analysis}
In this section, we aim to answer the following research questions:
\begin{itemize}

\item\textbf{(RQ1)} How can the underlying legal logic of the methods be interpreted? (Section \ref{section:A})

\item\textbf{(RQ2)} Is the text sequence of PLEs valid? (Section \ref{section:B})

\item\textbf{(RQ3)} What is the impact of the hyperparameter \(\lambda\) in the multi-task model? (Section \ref{section:C})
\end{itemize}

\subsection{Legal Logic Analysis}
\label{section:A}
As shown in Table~\ref{tab:threemodel} (a), to model the legal logic in the probation decision process, we establish three models with different frameworks and legal logic. In Table~\ref{tab:threemodel} (b), the results show that the multi-task model framework based on the {\it Dual-Track Theory of Punishment} performs the best.

\paragraph{TS-LE.}According to the two subtasks of the probation prediction task, in the TS-LE model, we use the fact description as input for Task 1. If the prediction result meets the prerequisites for probation eligibility, we use the text sequence of PLEs as input for Task 2 to predict the probation eligibility of criminals.

As shown in Table~\ref{tab:threemodel} (b), the TS-LE method achieves an accuracy of 95.17\% in Task 1. The legal logic behind this lies in Article 234 of the {\it Criminal Law of the People's Republic of China}, which specifies that in cases of intentional injury, sentencing is determined by the means employed by the criminal and the resulting consequences. Therefore, determining whether a criminal meets the prerequisites for probation relies on the fact description, which includes the criminal's actions and the outcomes. 

Next, we select cases from the dataset where the prediction result of Task 1 meets the prerequisites, and we input the text sequence of PLEs into the model to predict the probation eligibility of criminals. The results show that the accuracy is only 69.65\%, which is less than the accuracy of the MT-DT method (88.15\%). This reflects a legal logic where the independent training approach employed by the TS-LE method causes the two models trained in Task 1 and Task 2 to focus respectively on retributive and preventive factors. This approach fails to capture the dual retribution principle in \textit{Dual-Track Theory of Punishment}, which requires revisiting the retributive factors in Task 2.

\begin{figure}[!t]
    \centering
    \includegraphics[width=\columnwidth]{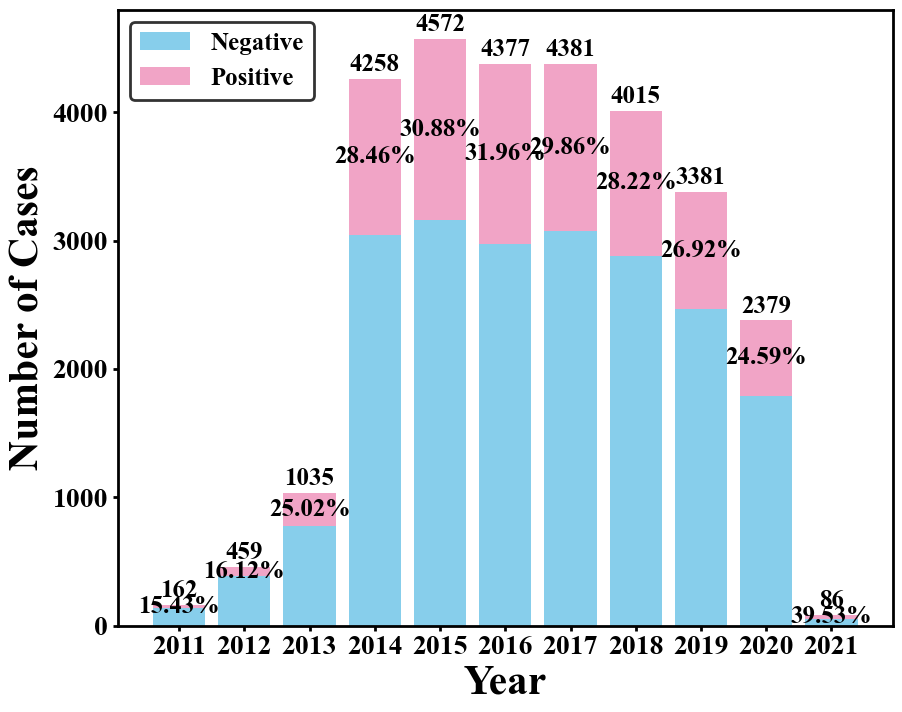} 
    \caption{Case distribution by year. From 2011 to 2021, probation accounted for only 28.69\% of all cases.}
    \label{fig:year}
\end{figure}

\begin{figure}[!t]  
\centering
\includegraphics[width=\linewidth]{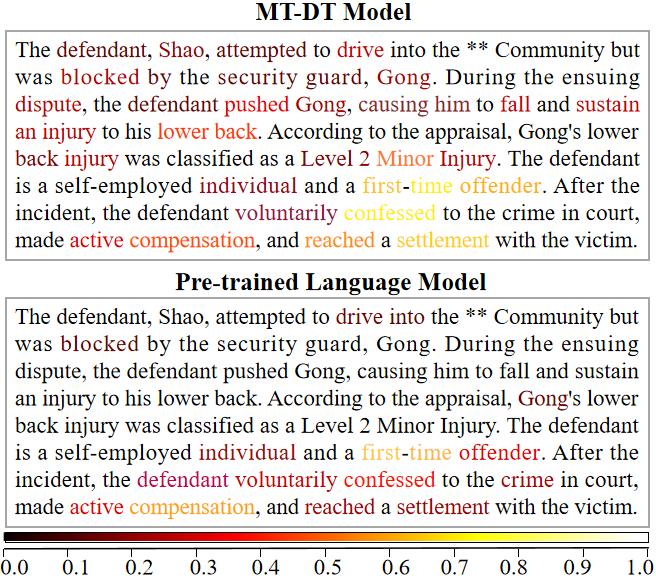}  
\caption{An example of attention score visualization of the MT-DT model and the pre-trained language model. The color changes from black to yellow, indicating that the model pays more and more attention to the text.}
\label{fig:att}
\end{figure}
\paragraph{TS-DT.}Based on the analysis in TS-LE, we propose a revised two-stage probation prediction method that aligns more closely with the \textit{Dual-Track Theory of Punishment} in legal logic. Specifically, in Task 2, we incorporate both the fact description and the text sequence of PLEs as joint inputs to the model for prediction. The experimental results in Table~\ref{tab:threemodel} (b) demonstrate that this model outperforms the TS-LE method in probation prediction. 

Furthermore, as shown in Figure~\ref{fig:year}, due to class imbalance in dataset, the cascade method used for training in Task 2 may lead to suboptimal model performance. The imbalance significantly hampers the model's ability to effectively predict probation eligibility. Moreover, since the prediction accuracy of Task 1 does not reach 100\%, any errors made in this task are likely to be amplified in Task 2. In practice, for cases where the true sentence is three years or less, but the Task 1 prediction incorrectly exceeds three years, these cases will not proceed to Task 2. Consequently, PLEs will not be considered and the probation eligibility cannot be evaluated. This results in disproportionately harsh sentencing outcomes for such criminals.

\paragraph{MT-DT.}To address the error amplification issue discussed in TS-DT, we adopt a joint training approach by modifying Task 1 to an implicit intermediate task. Thus, we propose a multi-task probation prediction model based on the \textit{Dual-Track Theory of Punishment}. This model treats Task 1 as an auxiliary task, jointly training Tasks 1 and 2.

In practice, we redefine the loss function of the model to ensure that, during training, the model simultaneously focuses on both the retributive and preventive factors of the criminal. According to the experimental results in Table~\ref{tab:threemodel} (b), the MT-DT method improves accuracy by 15.27\% compared to the TS-DT.

Additionally, we visualize the attention mechanism of the model, as shown in Figure~\ref{fig:att}. The visualization demonstrates that the model effectively captures both retributive and preventive factors in MT-DT. This indicates that our joint training framework effectively enhances the focus of the model on retributive factors, leading to better alignment with the legal logic of the \textit{Dual-Track Theory of Punishment}.

\begin{table}[!t]
  \centering
  \begin{minipage}[t]{\linewidth}
    \centering
    \captionsetup{labelformat=empty,skip=1pt}
    \setlength{\tabcolsep}{1.8mm}
    \resizebox{0.65\linewidth}{!}{
      \begin{tabular}{cccc}
      \toprule 
      \multirow{2}{*}{Input Information} & \multicolumn{3}{c}{Ablated Variants} \\
      \cline{2-4}
      & A & B & C \\
      \midrule
      Fact description & \ding{52} & \ding{52} & \ding{52} \\
      The vector of PLEs & \ding{55} & \ding{52} & \ding{55} \\
      The sequence of PLEs & \ding{55} & \ding{55} & \ding{52} \\
      \bottomrule        
      \end{tabular}
    }
    \caption*{(a) Different combinations of input information.}
  \end{minipage}
  
  
  \begin{minipage}[t]{\linewidth}
    \centering
    \captionsetup{labelformat=empty,skip=1pt}
    
    \setlength{\tabcolsep}{3mm}
    \resizebox{0.95\linewidth}{!}{
      \begin{tabular}{ccccc}
      \toprule 
      Ablated Variants & ACC(\%) & MP(\%) & MR(\%) & F1(\%) \\ \midrule
      A & 82.30 & 77.35 & 77.82 & 77.58 \\
      B & 83.42 & 78.90 & 80.85 & 79.75 \\
      C & 88.15 & 84.01 & 87.57 & 85.47 \\
      \bottomrule                                                                             
      \end{tabular}
    }
    \caption*{(b) Results with different input information}
  \end{minipage}
  
  \caption{Ablation study on input information}
  \label{tab:ablation}
\end{table}

\begin{table*}[!ht]
\centering
\setlength{\tabcolsep}{4mm}
\resizebox{0.85\linewidth}{!}{
    \begin{tabular}{cccccc}
    \toprule   
    & \multirow{1}{*}{Model} & \multirow{1}{*}{ACC(\%)} & \multirow{1}{*}{MP(\%)} & \multirow{1}{*}{MR(\%)} & \multirow{1}{*}{F1(\%)} \\  \midrule
    \multirow{5}{*}{\shortstack{Neural Network \\ Models}} 
    & TextCNN \cite{ref27} & 81.70 & 76.72 & 74.34 & 75.36\\ 
    & TextRNN \cite{ref28} & 81.56 & 77.30 & 72.01 & 73.87\\ 
    & Att-BLSTM \cite{ref29} & 81.19 & 76.00 & 78.93 & 77.13\\ 
    & TextRCNN \cite{ref30} & 82.29 & 77.28 & 76.31 & 76.77\\ 
    & FastText \cite{ref31} & 82.04 & 76.91 & 76.28 & 76.58\\ \midrule
    \multirow{5}{*}{\shortstack{Pre-trained \\ Language Models}} 
    & Lawformer \cite{ref21} & \underline{84.42} & 80.55 & 79.88 & 80.21\\ 
    & LagelBERT \cite{ref33} & 74.95 & 71.12 & 56.28 & 55.02\\ 
    & ELECTRA \cite{ref34} & 83.24 & 78.60 & 80.08 & 79.25\\ 
    & ALBERT \cite{ref35} & 82.40 & 77.85 & 76.11 & 76.89\\ 
    & BERT \cite{ref36} & 82.05 & \underline{82.34} & \underline{82.04} & \underline{82.01}\\ \midrule 
    & MT-DT (ours) & \textbf{88.15} & \textbf{84.01} & \textbf{87.57} & \textbf{85.47}\\ 
    \bottomrule 
    \end{tabular}}
    \caption{Evaluation of our model and baselines. The optimal performances are shown in bold font. The underlined values denote the optimal results of baselines.}
    \label{tab:model_performance}
\end{table*}
\subsection{Ablation Studies}
\label{section:B}
As shown in Table~\ref{tab:ablation} (a), we perform an ablation study on the MT-DT model. To examine the contribution of PLEs to performance, we consider the following three model variants:
\begin{itemize}
    \item \textbf{A.} Fact description only.
    \item \textbf{B.} Fact description and the vector of PLEs.
    \item \textbf{C.} Fact description and the text sequence of PLEs.
\end{itemize}

As shown in Table~\ref{tab:ablation} (b), replacing the vector with the text sequence of PLEs as input improves the accuracy by 4.73\%. Furthermore, the accuracy of both configurations involving PLEs (vector and text) exceeds that of using fact description alone. This suggests that the legal elements provide additional semantic information, thereby enhancing the model's ability to make accurate predictions.
\begin{figure}[!t]
    \centering
    \includegraphics[width=0.95\columnwidth]{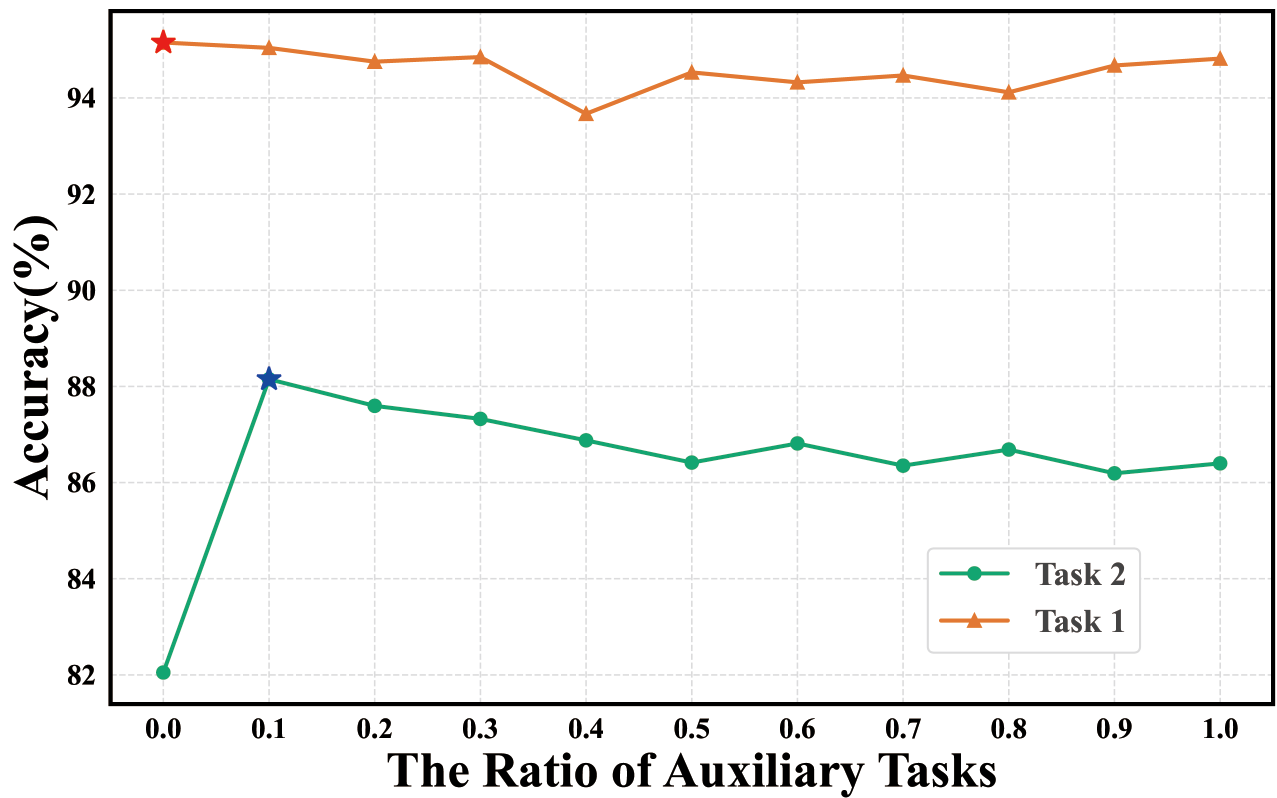}  
    \caption{Performance of the MT-DT model across different $\lambda$. The parameter is varied from 0 to 1 to assess its influence on the model's overall performance. The stars indicate maximum accuracy}
    \label{fig:lambda}
\end{figure}

\subsection{Influences of the Parameter \texorpdfstring{$\lambda$}{lambda}}

\label{section:C}
We further investigate the effect of the auxiliary task loss ratio in the overall loss on the performance of the MT-DT model. The results are shown in Figure~\ref{fig:lambda}. From the results, we observe the following:
\begin{itemize}
    \item When the auxiliary task loss is excluded from the loss function, the approach is comparable to directly using the pre-trained model for probation prediction. As a result, the model demonstrates suboptimal performance.  

    \item When the auxiliary task loss is incorporated, the performance of the MT-DT model stabilizes in the range of 86\% to 88\%, significantly surpassing the baseline. This demonstrates the robustness and effectiveness of the proposed method.
\end{itemize}

\section{Experiments}
\subsection{Dataset}
We obtain 29,105 judgment documents from publicly available legal documents on China judgements\footnote{\url{https://wenshu.court.gov.cn/}}. Of these, 8,351 cases involve probation, representing 28.69\% of the total. The dataset consists of first-instance intentional injury cases with a single defendant and a single charge, spanning from 2011 to 2021. The dataset was randomly split in an 8:1:1 ratio into training sets (23284 cases), validation sets (2911 cases), and test sets (2910 cases).

\subsection{Experimental Settings}
We set the encoder feature dimension at 768, with the maximum document length limited to 512 words. To prevent overfitting, dropout regularization was applied to the feature vectors. The batch size was set to 16. The models were fine-tuned using a batch contrastive loss function. For optimization, we employ the Adam optimizer \cite{ref37} with a learning rate of $10^{-5}$ and a dropout rate of 0.3. The model was trained for 10 epochs. The hyperparameters that performed the best, determined in the validation set, were evaluated in the test set, and the process was repeated six times to compute the average prediction results. The hyperparameters were set as follows: $\lambda = 0.1$.

For evaluation, we use the Accuracy (Acc), Macro Precision (MP), Macro Recall (MR), and Macro F1 (F1) metrics.


\subsection{Baseline}

In this section, we conduct a comprehensive comparison between our proposed MT-DT model and a range of conventional text classification models, as well as pre-trained language models widely used in natural language processing tasks. The detailed performance metrics for all models are summarized in Table~\ref{tab:model_performance}. The empirical results clearly demonstrate that MT-DT consistently outperforms all baseline models across various evaluation criteria, establishing its superior predictive capability in the context of the given tasks. Among the baseline models, LawFormer achieves the highest prediction accuracy, which is designed to effectively capture intricate patterns and perform sophisticated information extraction from large-scale, domain-specific legal corpora. However, despite its impressive accuracy, the complexity of the model leads to longer prediction times, which presents a challenge for real-time applications.

\section{Conclusion}
In this study, we propose a dataset for probation prediction, grounded in legal knowledge, and design an intelligent prediction model based on legal logic. By analyzing the substantive conditions for probation, we extract legal elements from legal documents and generate a text sequence, enriching the model with semantic information beyond the fact descriptions. We also introduce a novel loss function that enables the multi-task model to simultaneously focus on both facts and legal elements, thereby improving accuracy. Extensive experiments demonstrate that our method outperforms others and aligns with the legal logic of probation. Additionally, we assess the sensitivity of the model to parameters. 

In future research, we will conduct a deeper investigation into the key factors influencing probation sentence lengths, as well as systematically analyze the underlying legal principles governing such judicial determinations. Furthermore, we plan to extend our dataset to encompass cases involving a broader range of charges and incorporate more sophisticated legal reasoning frameworks to enhance the model’s robustness and generalizability.

\section*{Ethical Statement}

The dataset used in this study has been fully anonymized to protect the privacy of individuals. All data processing and analysis were conducted in compliance with relevant ethical guidelines and legal requirements.

\section*{Acknowledgements}
This work was supported by the National Natural Science Foundation of China under Grant T2293773 and Grant 12471488. Qinghua Wang and Xu Zhang contributed equally to this work. Please ask Dr. Cunquan Qu (cqqu@sdu.edu.cn) for correspondence.

\bibliographystyle{named}
\bibliography{main.bbl}

\end{document}